\documentclass[runningheads]{llncs}
\usepackage{graphicx}
\graphicspath{ {./figures/} }
\usepackage{multirow}

\usepackage[misc]{ifsym}
 
\usepackage{makecell}

\begin{document}
\title{Sentence Constituent-Aware Aspect-Category Sentiment Analysis with Graph Attention Networks}
\titlerunning{Sentence Constituent-Aware Aspect-Category Sentiment Analysis}
%
\author{Yuncong Li\inst{1}\textsuperscript{($\star$)} \and
	Cunxiang Yin\inst{1}\textsuperscript{($\star$)} \and
	Sheng-hua Zhong\inst{2}\textsuperscript{($\dagger$)}}
\authorrunning{Y. Li et al.}
%
\institute{	Baidu Inc., Beijing, China\\
	\email{\{liyuncong,yincunxiang\}@baidu.com} \and
	College of Computer Science and Software Engineering, Shenzhen University, Shenzhen, China\\
	\email{csshzhong@szu.edu.cn}}
\maketitle              

\renewcommand{\thefootnote}{\fnsymbol{footnote}}
\footnotetext[1]{Equal contribution.}
\footnotetext[4]{Corresponding author.}

\renewcommand{\thefootnote}{\arabic{footnote}}

\begin{abstract}
Aspect category sentiment analysis (ACSA) aims to predict the sentiment polarities of the aspect categories discussed in sentences. Since a sentence usually discusses one or more aspect categories and expresses different sentiments toward them, various attention-based methods have been developed to allocate the appropriate sentiment words for the given aspect category and obtain promising results. However, most of these methods directly use the given aspect category to find the aspect category-related sentiment words, which may cause mismatching between the sentiment words and the aspect categories when an unrelated sentiment word is semantically meaningful for the given aspect category. To mitigate this problem, we propose a Sentence Constituent-Aware Network (SCAN) for aspect-category sentiment analysis. SCAN contains two graph attention modules and an interactive loss function. The graph attention modules generate representations of the nodes in sentence constituency parse trees for the aspect category detection (ACD) task and the ACSA task, respectively. ACD aims to detect aspect categories discussed in sentences and is a auxiliary task. For a given aspect category, the interactive loss function helps the ACD task to find the nodes which can predict the aspect category but can’t predict other aspect categories. The sentiment words in the nodes then are used to predict the sentiment polarity of the aspect category by the ACSA task. The experimental results on five public datasets demonstrate the effectiveness of SCAN. \footnote{Data and code can be found at https://github.com/l294265421/SCAN}

\end{abstract}
\section{Introduction}
\begin{figure}
	\centering
	\includegraphics[width=0.94\textwidth]{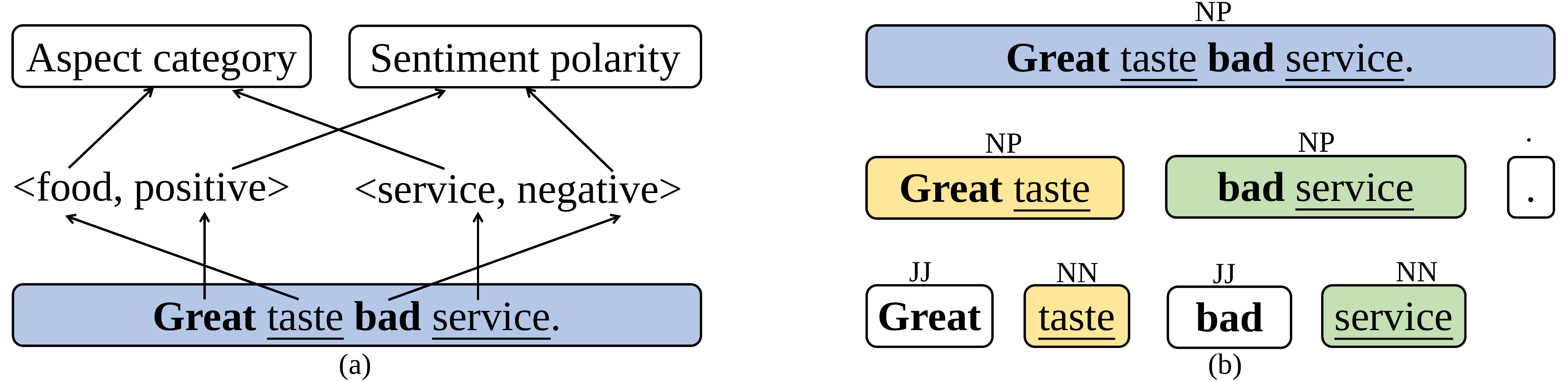}
	\caption{(a) An example of aspect-category sentiment analysis. (b) The constituency parse tree of the sentence in (a) generated by the Berkeley Neural Parser \cite{Kitaev-2018-SelfAttentive}. The underlined words are aspect terms and the bold words are sentiment words.
	} \label{example}
\end{figure}

Aspect based sentiment analysis (ABSA) \cite{pontiki2016semeval,pontiki2015semeval,pontiki-etal-2014-semeval} is a fine-grained sentiment analysis task. ABSA contains several subtasks, four of which are aspect category detection (ACD) detecting aspect categories mentioned in sentences, aspect category sentiment analysis (ACSA) predicting the sentiments of the detected aspect categories, aspect term extraction (ATE) identifying aspect terms presenting in sentences and aspect term sentiment analysis (ATSA) classifying the sentiments toward the identified aspect terms. While aspect categories mentioned in a sentence are from a few predefined categories and may not occur in the sentence, aspect terms  explicitly appear in sentences. Fig.~\ref{example} (a) shows an example. ACD detects the two aspect categories \emph{food} and \emph{service} and ACSA predicts the positive and negative sentiments toward them. ATE identifies the two aspect terms ``taste'' and ``service'' and ATSA classifies the positive and negative sentiments toward them. In this paper, we concentrate on the ACSA task. The ACD task as a auxiliary is used to find aspect category-related nodes from sentence constituency parse trees for the ACSA task.

Since a sentence usually discusses one or more aspect categories and expresses different sentiments toward them, various attention-based methods have been developed to allocate appropriate sentiment words for given aspect categories. Wang et al. \cite{wang2016attention} were the first to explore attention mechanism on the ACSA task and proposed an attention based LSTM (AT-LSTM). For a given sentence and an aspect category mentioned in the sentence, AT-LSTM first models the sentence via a LSTM model,  then combines the hidden states from the LSTM with the representation of the aspect category to generate aspect category-specific word representations, finally applies an attention mechanism over the word representations to find the aspect category-related sentiment words, that are used to predict the sentiment of the aspect category. The constrained attention networks (CAN) \cite{hu2019can} handles multiple aspect categories of a sentence simultaneously and introduces orthogonal and sparse regularizations to constrain the attention weight allocation. The aspect-level sentiment capsules model(AS-Capsules) \cite{wang2019aspect} performs ACD and ACSA simultaneously, which also uses an attention mechanism to find aspect category related sentiment words and achieves state-of-the-art performances on the ACSA task.

However, these models directly use the given aspect category to find the aspect category-related sentiment words, which may cause mismatching between the sentiment words and the aspect categories when an unrelated sentiment word is semantically meaningful for the given aspect category. For the example in Fig.~\ref{example}, ``Great'' and ``bad'' can be used interchangeably. It is hard for attention-based methods to distinguish which word is associated with aspect category \emph{food} or \emph{service} among ``good'' and ``bad''. To solve the problem, The HiErarchical ATtention(HEAT) network \cite{cheng2017aspect} first finds the aspect terms indicating the given aspect cagegory, then finds the aspect category-related sentiment words  depending on the position information and semantics of the aspect terms. Although HEAT obtains good results, to train HEAT, we additionally need to annotate the aspect terms indicating the given aspect category, which can be time-consuming and expensive.

To mitigate the mismatch problem, we propose a Sentence Constituent-Aware Network (SCAN) for aspect-category sentiment analysis which does not require any additional annotation. SCAN contains two graph attention networks (i.e. GAT for ACD and GAT for ACSA) \cite{velivckovic2017graph} and an interactive loss function. Given a sentence, we first use the Berkeley Neural Parser \cite{Kitaev-2018-SelfAttentive} to generate the constituency parse tree. The two GATs generate representations of the nodes in the sentence constituency parse tree for the ACD task and the ACSA task, respectively. The GAT for ACD mainly attends to the words indicating aspect categories, while the GAT for ACSA mainly attends to sentiment words. For a given aspect category, the interactive loss function helps the ACD task to find the nodes that can predict the aspect category but can’t predict other aspect categories. The sentiment words in the nodes then are used to predict the sentiment polarity of the aspect category by the ACSA task. Fig.~\ref{example} (b) shows the constituency parse tree of the sentence ``Greate taste bad service.''. For the aspect category \emph{food}, SCAN first finds the yellow nodes ``Greate taste'' and ``taste'', then predict the sentiment of \emph{food} based on the sentiment word ``Great'' in the node ``Great taste''. SCAN excludes the blue node ``Great taste bad service.'' for \emph{food}, because it can predict not only \emph{food} but also \emph{service}.

The main contributions of our work can be summarized as follows:
\begin{itemize}
	\item We propose a Sentence Constituent-Aware Network (SCAN) for aspect-category sentiment analysis, which generates representations of the nodes in constituency parse trees with graph attention networks.
	\item We introduce an interactive loss function, helping SCAN to exclude the nodes in constituency parse trees that can also predict other aspect categories for the given aspect category.
	\item The experimental results on five public datasets demonstrate the effectiveness of SCAN.
\end{itemize}

\section{Related Work}
\subsection{Aspect-Category Sentiment Analysis}
Aspect-category sentiment analysis (ACSA) aims to predict the sentiments of a sentence toward the given aspect categories. We summarize previous approaches for this task into two classes: non-attention based models and attention-based models. Given a sentence and an aspect category mentioned in the sentence, non-attention based models \cite{sun2019utilizing,xue2018aspect} directly generate the aspect category-specific sentence representation, then predict the sentiment of the aspect category based on the representation. Although some non-attention based models (e.g. BERT-pair-QA-B \cite{sun2019utilizing}) achieve state-of-the-art results, they don't provide reasons why they make a prediction, so, they lack interpretability.

Compared with non-attention based models, attention based models \cite{hu2019can,jiang2019challenge,tay2018learning,wang2016attention,wang2019aspect,10.1145/3350487} are more interpretable. They first find aspect category-related sentiment words, then generate aspect category-specific representations based on the sentiment words. Attention mechanism was first used by Wang et al. \cite{wang2016attention} to find aspect category-related sentiment words. Jiang et al. \cite{jiang2019challenge} proposed new capsule networks (CapsNet and CapsNet-BERT) to model the complicated relationship between aspect categories and context, where the normalization weights and routing weights can be viewed as attention weights. CAN \cite{hu2019can} and AS-Capsules \cite{wang2019aspect} perform the ACD task and the ACSA task jointly and achieves state-of-the-art performances. However, these attention based models directly use the given aspect category to find the aspect category-related sentiment words, which may cause the mismatch problem mentioned above.

\subsection{Syntax-aware Aspect-Term Sentiment Analysis}
Since SCAN is, to the best of our knowledge, the first syntax-aware model for ACSA, we review some syntax-aware methods for the Aspect-Term Sentiment Analysis (ATSA) task. ATSA predicts the sentiments of the given aspect terms that occur in a sentence. Early methods for ATSC try to incorporate syntax knowledge using recursive neural networks. Dong et al. \cite{dong-etal-2014-adaptive} first converted the dependency tree of a sentence to a binary tree. They proposed an Adaptive Recursive Neural Network (AdaRNN) that is applied on the binary tree to propagate the sentiments of words to the target node. The representation of the target node is used to predict the sentiment label of the target. Similar to aspect term, the target is a sequence of words that occur in the sentence. Phrase Recursive Neural Network (PhraseRNN) \cite{nguyen-shirai-2015-phrasernn} makes the representation of the aspect term richer by using syntactic information from both the dependency and constituent trees of sentences. These methods that have to convert the original dependency tree into a binary tree may move modifying sentiment words farther away from the aspect term. Recently, A few researches use graph neural network \cite{4700287} to incorporate syntax knowledge and achieve the state-of-the-art performance. Huang et al. \cite{huang-carley-2019-syntax} applied graph attention network (GAT) \cite{velivckovic2017graph} over the dependency tree of a sentence and the representations of the nodes corresponding to aspect terms is used to predict the sentiment of the aspect terms. Zhang et al. \cite{zhang-etal-2019-aspect} applied graph convolutional network (GCN) over the dependency tree of a sentence, then the representation of the node corresponding to the given aspect term is used to retrive aspect-related sentiment words that are used to predict the sentiment of the aspect term. Since aspect categories may not occur in sentences, these methods can't be used for the ACSA task directly.

\section{Model}
\begin{figure}
	\centering
	\includegraphics[width=0.99\textwidth]{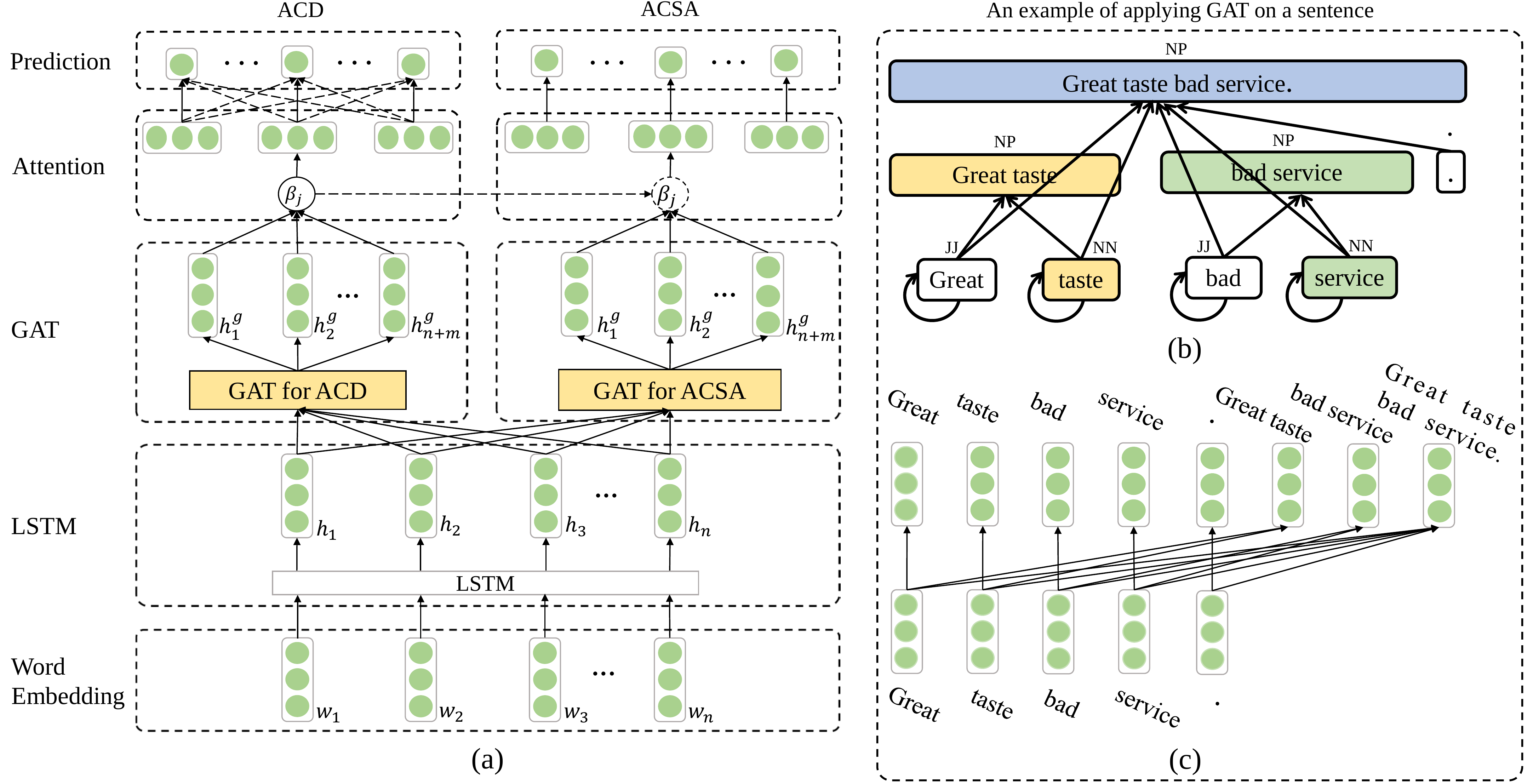}
	\caption{(a) The overall architecture of SCAN. (b) An example of generating the graph of a constituency parse tree. (c) An example of applying GAT on the graph.
	} \label{SCAN}
\end{figure}

We first formulate the problem. There are $N$ predefined aspect categories, $A=\{A_1,A_2,...,A_N\}$, and $M$ sentiment labels, $P=\{P_1,P_2,...,P_M\}$. Given a sentence, $S=\{w_1,w_2,...,w_n\}$, SCAN identifies the $K$ aspect categories mentioned in the sentence, $A^S=\{A_1^S,A_2^S,...,A_K^S\}$, $A^S \subset A$, and predicts the sentiments toward them, $P^S=\{P_1^S,P_2^S,...,P_K^S\}$.
The overall architecture of SCAN is shown in Fig.~\ref{SCAN} (a).  The modules of SCAN will be introduced in the rest of this section.

\subsection{Embedding Layer}
In the embedding layer, we convert the sentence $S$ into word embeddings $E=\{e_1,e_2,...,e_n\}$, where $e_i \in R^d$ and d is the embedding dimension. 

\subsection{LSTM Layer}
The word embeddings of the sentence are then fed into a LSTM network \cite{hochreiter1997long}, which outputs hidden states $H = \{h_1,h_2,...,h_n\}$. The size of the hidden state is also set to be d. 

\subsection{Graph Attention over Sentence Constituency Parse Trees}
A graph attention network (GAT) \cite{velivckovic2017graph} is a variant of graph neural network \cite{4700287}. We apply GAT over the constituency parse trees of sentences. For the given sentence, we use the Berkeley Neural Parser \cite{Kitaev-2018-SelfAttentive} to generate the constituency parse tree, then construct a directed graph based on the tree. The graph contains $n+m$ nodes, where $n$ nodes are from the leaf nodes and $m$ nodes from the internal nodes. In the graph, there are edges between the leaf nodes and themselves as well as their ancestor nodes. The leaves are source nodes. For the graph, where each leaf node is associated with corresponding hidden state from the LSTM layer, one GAT layer updates node representations by aggregating neighbourhood’s hidden states. Fig. \ref{SCAN} (b) and (c) show an example of applying GAT on a sentence.

Specifically, given a node $i$ with a hidden state $h_i$ and its neighbours $n[i]$ as well as their hidden states, a  GAT layer updates the node’s hidden state using multi-head attentions \cite{vaswani2017attention}. The new hidden representation $h^g_i$ of the node $i$ is computed by:
\begin{equation}
h^g_i=\left\|_{l=1}^{L}\right.\sigma(\sum_{j \in n[i]}\alpha^{ij}_lW_lh_j)
\end{equation}
\begin{equation}
\label{gat-e}
\alpha^{ij}_l=\frac{exp(f(a^T_l[W_lh_i\|W_lh_j]))}{\sum_{u \in n[i]}exp(f(a^T_l[W_lh_i\|W_lh_u]))}
\end{equation}
where $\left\|\right.$ represents vector concatenation, $\alpha^{ij}_l$ is the attention coefficient of node $i$ to its neighbour $j$ in attention head $l$. $W_l \in R^{\frac{d}{L} \times d}$ is a linear transformation matrix for input states. $\sigma$ denotes a sigmoid function. $f(.)$ is a LeakyReLU non-linear function. $a_l \in R^{\frac{2d}{L}}$ is an attention context vector learned during training. When node $i$ is a internal node, its hidden state $h_i$ is a zero vector.

For simplicity, we can write such feature propagation process as
\begin{equation}
H^g=GAT(H,A;\Theta^g)
\end{equation}
where $H^g \in R^{(n+m) \times d}$ is the stacked states for all nodes, $A \in R^{(n+m) \times (n+m)}$ is the graph adjacent matrix. $\Theta^g$ is the parameter set of the GAT.

We apply two different GATs on the graph for ACD and ACSA, respectively:
\begin{equation}
H^g_{ACD}=GAT_{ACD}(H,A;\Theta^g_{ACD})
\end{equation}
\begin{equation}
H^g_{ACSA}=GAT_{ACSA}(H,A;\Theta^g_{ACSA})
\end{equation}

\subsection{Attention Layer}
 This layer takes $H^g_{ACD}$ as input, and produces attention \cite{yang2016hierarchical} weight vectors for all predefined aspect categories. Formally, for the $j$-th aspect category:
 \begin{equation}
 M_j=tanh(W_jH^g_{ACD}+b_j)
 \end{equation}
 \begin{equation}
 \label{attention-e}
\beta_j=softmax(u^T_jM_j)
 \end{equation}
where $W_j \in R^{d \times d}$, $b_j \in R^d$, $u_j \in R^d$ are learnable parameters, and $\beta_j$ is the attention weight vector.

\subsection{Prediction Layer}
The attention weights are applied on $H^g_{ACD}$ to generate aspect category-specific sentence representations for the ACD task. For the $j$-th aspect category:
 \begin{equation}
\hat{r}_j^{ACD}=H^g_{ACD}\beta_j^T
\end{equation}

The representation of the $j$-th aspect category then is used to predict all predifined aspect categories:
 \begin{equation}
\hat{y}_{j_i}^{ACD}=sigmoid(W_i\hat{r}_j^{ACD}+b_i),i=1,2,..,N
\end{equation}
where $W_i \in R ^{d \times 1}$ and $b_i \in R$.

The attention weights are applied on $H^g_{ACSA}$ to generate aspect category-specific sentence representations for the ACSA task. For the $j$-th aspect category:
\begin{equation}
\hat{y}_j^{ACSA}=softmax(W^2ReLU(W^1(H^g_{ACSA}\beta_j^T)+b_j^1)+b_j^2)
\end{equation}
where $W^1 \in R^{d \times d}$, $W^2 \in R^{d \times M}$, $b^1 \in R^d$ and $b^2 \in R^ M$.

\subsection{Loss}
For the ACD task, the main loss function is defined by:
\begin{equation}
L_{ACD}=-\sum_{j=1}^{N}y_{j_j}^{ACD}log\hat{y}_{j_j}^{ACD}+(1-y_{j_j}^{ACD})log(1-\hat{y}_{j_j}^{ACD})
\end{equation}

The interactive loss (iLoss) we propose for the ACD task is defined by:
\begin{equation}
L_{iLoss}=-\frac{1}{N-1}\sum_{j=1}^{N}\sum_{i=1,i \neq j}^{N}log(1-\hat{y}_{j_i}^{ACD})
\end{equation}
where $log(1-\hat{y}_{j_i}^{ACD})$ is a special case of $y_{j_i}^{ACD}log\hat{y}_{j_i}^{ACD}+(1-y_{j_i}^{ACD})log(1-\hat{y}_{j_i}^{ACD})$ where $y_{j_i}^{ACD}=0$. The iLoss punishes the SCAN when the representation of a particular aspect category can predict other aspect categories. 

For the ACSA task, the loss function is defined by:
\begin{equation}
L_{ACSA}=-\sum_{j=1}^{K}\sum_{c \in P}y_{j_c}^{ACSA}log\hat{y}_{j_c}^{ACSA}
\end{equation}

The parameters are trained by minimizing the combined loss function:
\begin{equation}
L(\theta)=\epsilon L_{ACD}+\eta L_{iLoss}+\mu L_{ACSA}+\lambda \left \| \theta \right \|^2_2
\end{equation}
where $\epsilon$, $\eta$ and $\mu$ is the weights of the three losses, respectively. $\lambda$ is the $L2$ regularization factor and $\theta$ contains all parameters of SCAN.

\section{Experiments}
\subsection{Datasets}
We conduct experiments on five datasets. \textbf{Rest14} is the restaurant review dataset of SemEval 2014 Task 4 \cite{pontiki-etal-2014-semeval}. Following \cite{xue2018aspect}, \textbf{RestLarge} is obtained by merging the restaurant review datasets of SemEval 2014 Task 4, 
SemEval-2015 Task 12 \cite{pontiki2015semeval}, and SemEval-2016 Task 5 \cite{pontiki2016semeval}. Incompatibilities of data are fixed during merging. After aspect categories are merged, if an aspect category has both positive and negative polarities, it is assigned conflict polarity. we remove
samples with conflict polarities from Rest14 and RestLarge. Since most sentences in Rest14 and RestLarge contain only one aspect category or multiple aspect categories with the same sentiment polarity, we construct \textbf{Rest14-hard} and \textbf{RestLarge-hard} to measure the ability of models detecting multiple different sentiment
polarities of one sentence toward different aspect categories. The training set and development set of Rest14-hard (RestLarge-hard) are the same as Rest14's (RestLarge's), and the test set of Rest14-hard (RestLarge-hard) only includes the sentences in Rest14's (RestLarge's) test set containing at least two aspect categories with different sentiment polarities. \textbf{MAMS-ACSA} is released by Jiang et al. \cite{jiang2019challenge}, all sentences in which contain multiple aspect categories with different sentiment polarities. Statistics of these datasets are given in Table~\ref{datasets}.

\begin{table}
	\centering
	\caption{Statistics of the datasets.}
	\label{datasets}
	\begin{tabular}{ |c|c|c|c|c|c|c|c|c|c|c|c| } 
		\hline
		\multirow{2}*{Polarity} & \multicolumn{3}{|c|}{Rest14} & Rest14-hard & \multicolumn{3}{|c|}{RestLarge} & RestLarge-hard & \multicolumn{3}{|c|}{MAMS-ACSA} \\ 
		\cline{2-12}
		 & Train & Dev & Test & Test & Train & Dev & Test & Test & Train & Dev & Test \\ 
		 \hline
		 Pos. & 855 & 324 & 657 & 21 & 2550 & 646 & 1553 & 90 & 1929 & 241 & 245 \\ 
		 \hline
		 Neg. & 733 & 106 & 222 & 20 & 928 & 242 & 710 & 87 & 2084 & 259 & 263 \\ 
		 \hline
		 Neu. & 430 & 70 & 94 & 12 & 438 & 110 & 172 & 45 & 3077 & 388 & 393 \\ 
		 \hline
	\end{tabular}
	
\end{table}

\subsection{Implementation Details}
We implement all models in Pytorch \cite{paszke2017automatic}. We use 300-dimentional word vectors pre-trained by GloVe \cite{pennington2014glove} to initialize the word embedding vectors. The batch sizes are set to 32 for non-BERT models and 16 for BERT-based models, respectively. All models are optimized by the Adam optimizer \cite{kingma2014adam}. The learning rates are set to 0.001 and 0.00002 for non-BERT models and BERT-based models, respectively. We set $L=4$, $\epsilon=1$, $\eta=1$, $\mu=1$, and $\lambda=0.00001$. For SCAN-BERT, we first train SCAN to obtain the nodes weights $\beta_j$ for each aspect category, SCAN-BERT then uses these weights to find aspect category-related nodes and set $\epsilon=0$ as well as $\eta=0$. That is, SCAN-BERT only improves the ability of SCAN that predicts the sentiments based on aspect category-specific representations. We apply early stopping in training and the patience is 10. We run all models for 5 times and report the average results on the test datasets.

\subsection{Compared Methods}
We compare SCAN with various baselines. (1) sentence-level sentiment analysis models: LSTM \cite{jiang2019challenge}, TextCNN \cite{jiang2019challenge}, BilstmAttn \cite{jiang2019challenge} and BERT-single \cite{devlin2019bert};(2) non-BERT ACSA models: AT-LSTM \cite{wang2016attention}, ATAE-LSTM \cite{wang2016attention}, GCAE \cite{xue2018aspect} and CapsNet \cite{jiang2019challenge}; (3) non-BERT joint models that perform the ACD task and the ACSA task simutaniously: As-capsule \cite{wang2019aspect} and M-AT-LSTM \cite{hu2019can}; (4) BERT based models: BERT-pair \cite{devlin2019bert}, BERT-pair-QA-B \cite{sun2019utilizing} and CapsNet-BERT \cite{jiang2019challenge}. We also provide the comparisons of several variants of SCAN:

\textbf{SCAN -w/o iLoss} removes the iLoss of SCAN.

\textbf{SCAN -w/o tree} removes the GAT layer of SCAN.

\textbf{SCAN-BERT} replaces the embedding layer and the LSTM layer in SCAN with the uncased basic pre-trained BERT.

\textbf{SCAN-BERT-AVE} replaces the GAT for ACSA with an average GAT in SCAN-BERT. The average GAT updates a node’s hidden state with the average of the hidden states of its neighbors.

\begin{table}
	\centering
	\caption{Results of the ACSA task in terms of accuracy (\%). $\dagger$ refers to citing from CapsNet \cite{jiang2019challenge}. Best scores are marked in bold.}
	\label{experimental-results}
	\begin{tabular}{ l|p{4em}<{\centering}|c|c|c|c } 
		\hline
		Method & Rest14 & Rest14-hard & RestLarge & RestLarge-hard & MAMS-ACSA\\
		\hline
		LSTM & 80.905 & 47.924 & 78.483 & 45.373 & 46.614\\
		TextCNN & \textbf{83.556} & 51.698 & 78.235 & 46.790 & 49.056\\
		BilstmAttn & 82.158 & 50.189 & 77.994 & 46.119 & 48.568\\
		AT-LSTM & 82.672 & 54.717 & 77.609 & 50.298 & 66.436$\dagger$\\
		ATAE-LSTM & 82.138 & 56.604 & 78.363 & 52.686 & 70.634$\dagger$\\
		GCAE & 81.336 & 54.717 & 77.841 & 51.343 & 72.098$\dagger$\\
		CapsNet & 81.172 & 53.962 & 79.859 & 56.577 & 73.986$\dagger$\\
		M-AT-LSTM & 81.275 & 60.755 & 80.240 & 56.967 & 74.650\\
		As-capsule & 82.179 & 60.755 & 80.678 & 63.688 & 75.116\\
		SCAN & 80.699 & \textbf{68.302} & 80.315 & \textbf{67.541} & 75.405\\
		SCAN -w/o iLoss & 80.370 & 65.660 & \textbf{80.960} & 65.410 & 74.828\\
		SCAN -w/o tree & 80.123 & 62.264 & 80.182 & 63.197 & \textbf{76.582}\\
		\hline
		BERT-single & 87.112 & 49.434 & 82.430 & 46.940 & 47.635\\
		BERT-pair & 87.482 & 67.547 & 84.059 & 60.149 & 78.292$\dagger$\\
		BERT-pair-QA-B & 87.523 & 69.433 & 86.686 & 70.862 & 79.134\\
		CapsNet-BERT & 86.557 & 51.321 & 86.040 & 55.766 & 79.461$\dagger$\\
		SCAN-BERT & \textbf{88.942}  & 70.189 & 86.595 & 69.672 & 79.883\\
		SCAN-BERT-AVE & 88.613  & \textbf{70.943} & \textbf{87.009} & \textbf{71.967} & \textbf{80.444}\\
		\hline
	\end{tabular}
\end{table}

\subsection{Results}
Table~\ref{experimental-results} shows our experimental results on the five datasets. From
Table~\ref{experimental-results} we draw the following conclusions. First, SCAN outperforms all non-bert baselines on the Rest14-hard dataset, the RestLarge-hard dataset and the MAMS-ACSA dataset, which shows that SCAN can better recognize the different sentiment polarities of sentences toward different aspect categories. Second, SCAN outperforms SCAN -w/o iLoss in 4 of 5 results and also outperforms SCAN -w/o tree in 4 of 5 results, which indicates that both the proposed interactive loss and applying graph attention over sentence constituency parse trees can improve the performance of the ACSA task. Third, SCAN-BERT-AVE surpasses all other baselines on all five datasets, indicating that the aspect category-related nodes in sentence constituency parse trees found by SCAN can be used to effectively predict the sentiment polarities of aspect categories. Fourth, SCAN-BERT obtains worse results than SCAN-BERT-AVE, the possible reason is that the GAT for ACSA is not fully trained during fine-tuning BERT. In addition, TextCNN surpasses all non-BERT ACSA models on the Rest14 dataset, which indicates that Rest14 is not effective enough to be used to  evaluate the ability of models to address the specific challenge of ACSA: detect multiple different sentiment
polarities of one sentence toward different aspect categories.

\begin{figure}
	\centering
	\includegraphics[width=0.94\textwidth]{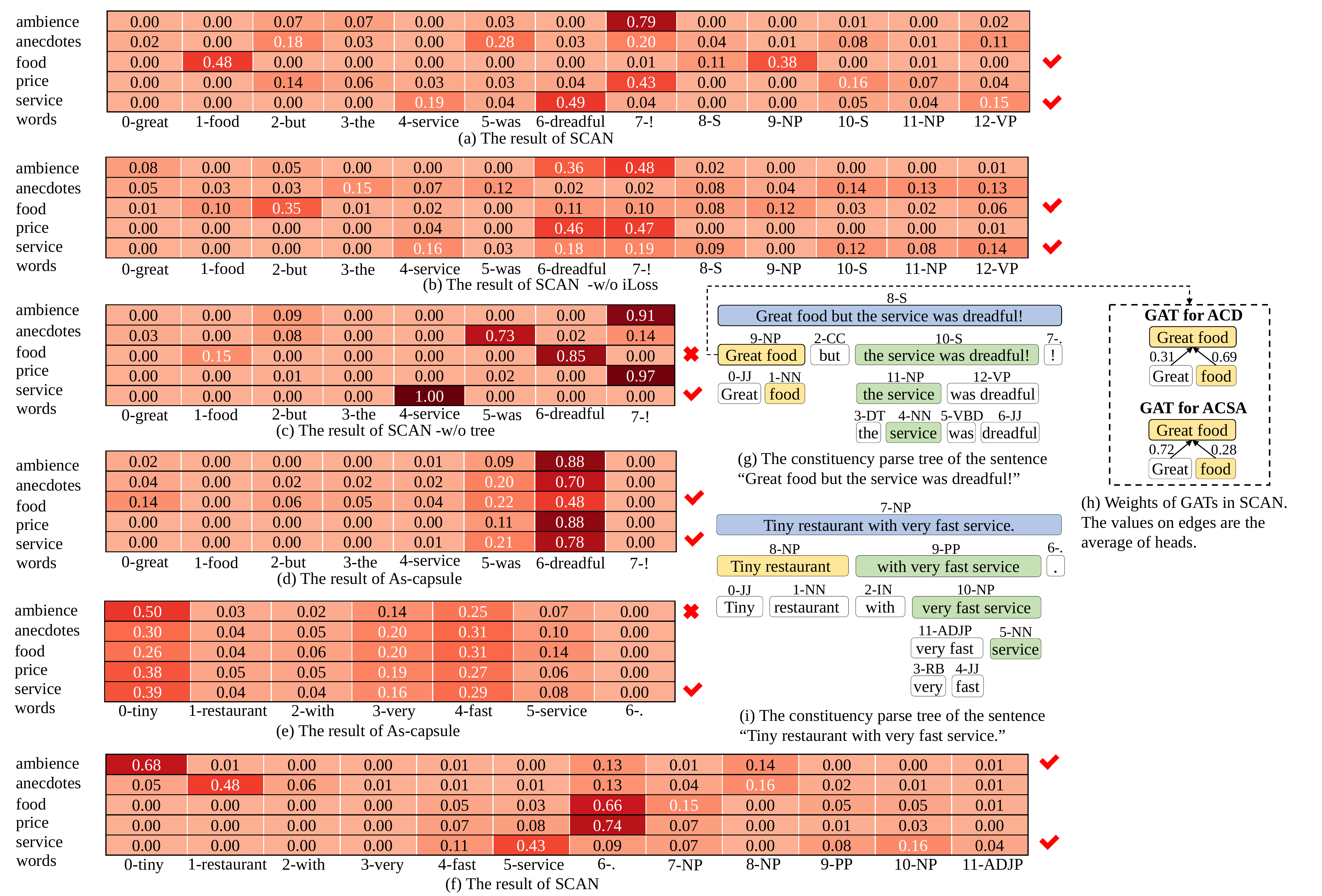}
	\caption{Visualization of attention weights.} \label{visualization-of-attention-weights}
\end{figure}

\subsection{Attention Visualizations}
To show that our models can alleviate the mismatch problem, Fig.~\ref{visualization-of-attention-weights} visualizes the attention weight $\beta_j$ in Equation \ref{attention-e} and $\alpha^{ij}_l$ in Equation \ref{gat-e} of our models. We also visualize the sentiment attention weights of As-capsule for comparison. Fig.~\ref{visualization-of-attention-weights} (a) shows that SCAN accurately finds the nodes ``food'' and ``Great food'' for the aspect category \emph{food}, and finds the nodes ``service'', ``dreadfull'', ``the service was dreadful'', ``the service'' and ``was dreadful'' for \emph{service}. Fig.~\ref{visualization-of-attention-weights} (h) shows that, for the node ``Great food'', the GAT for ACD mainly attends to ``food'' indicating \emph{food} and the GAT for ACSA mainly attends to the sentiment word ``Great''. Fig.~\ref{visualization-of-attention-weights} (b) and (c) show that both SCAN  -w/o iLoss and SCAN -w/o tree wrongly find the node ``dreadful'' for \emph{food} and Fig.~\ref{visualization-of-attention-weights} (d) shows that As-capsule also wrongly finds the word ``dreadful'' for \emph{food}, which indicate that the interactive loss and applying graph attention over sentence constituency parse trees form a complete solution to alleviate the mismatch problem, as any one of them alone does not address the issue. We also observe that, in Fig.~\ref{visualization-of-attention-weights} (a) and (c), the attention weights of the aspect categories mentioned in the sentence is nearly orthogonal and the attention of the aspect categories that don't be mentioned in the sentence mainly attend to the meaningless stop words. It indicates that our proposed interactive loss has similar effect to the orthogonal regularization proposed by \cite{hu2019can}. However, to use the orthogonal regularization, sentences must have extra annotation information that whether texts describing different aspect categories overlap. Fig.~\ref{visualization-of-attention-weights} (e) shows that As-capsule wrongly predict the sentiment polarity of \emph{ambience} in the sentence ``Tiny restaurant with very fast service.'' because of the mismatch problem while Fig.~\ref{visualization-of-attention-weights} (f) shows that SCAN correctly finds the related nodes and predicts the sentiment polarity of \emph{ambience}. 

\section{Conclusion}
In this paper, We propose a Sentence Constituent-Aware Network (SCAN) for aspect-category sentiment analysis. The two graph attention modules and the interactive loss function in SCAN form a complete solution to alleviate the mismatch problem. The experimental results on five public datasets demonstrate the effectiveness of SCAN. Future work could consider making the representations of the leaf nodes richer by using syntactic information from the dependency tree of the sentence and modelling the inter-aspect category dependencies.
%
%
%
\bibliographystyle{splncs04}
\bibliography{nlpcc}

\end{document}